\newcommand{\CLASSINPUTinnersidemargin}{0.67in} % left
\newcommand{\CLASSINPUToutersidemargin}{0.67in} % right
\newcommand{\CLASSINPUTtoptextmargin}{0.85in}   % top
\newcommand{\CLASSINPUTbottomtextmargin}{1.05in}% bottom
\documentclass[conference]{IEEEtran}
\IEEEoverridecommandlockouts
\usepackage{cite}
\usepackage{amsmath,amssymb,amsfonts}
\usepackage{graphicx}
\usepackage{textcomp}
\usepackage{xcolor}
\usepackage{algorithm}
\usepackage{tabularx}
\usepackage{algpseudocode}   % from algorithmicx
\usepackage{setspace}        % only if you want increased line spacing
\usepackage{booktabs}  % nicer table rules
\usepackage{url} % for \url command
\usepackage{array}
\usepackage{siunitx}
\usepackage{balance}
\usepackage{subcaption}
\def\BibTeX{{\rm B\kern-.05em{\sc i\kern-.025em b}\kern-.08em
    T\kern-.1667em\lower.7ex\hbox{E}\kern-.125emX}}
\begin{document}

\title{HeatPrompt: Zero-Shot Vision-Language Modeling of Urban Heat Demand from Satellite Images\\
\thanks{*corresponding author kundan.thota@kit.edu}
}

\author{
Kundan Thota, Xuanhao Mu, Thorsten Schlachter, Veit Hagenmeyer\\

Institute for Automation and Applied Informatics (IAI), Karlsruhe Institute of Technology (KIT), Germany \\
\{kundan.thota, xuanhao.mu, thorsten.schlachter, veit.hagenmeyer\}@kit.edu
}

\maketitle

\begin{abstract}
Accurate heat-demand maps play a crucial role in decarbonizing space heating, yet most municipalities lack detailed building-level data needed to calculate them. We introduce HeatPrompt, a zero-shot vision-language energy modeling framework that estimates annual heat demand using semantic features extracted from satellite images, basic Geographic Information System (GIS), and building-level features. We feed pretrained Large Vision Language Models (VLMs) with a domain-specific prompt to act as an energy planner and extract the visual attributes such as roof age, building density, etc, from the RGB satellite image that correspond to the thermal load. A Multi-Layer Perceptron (MLP) regressor trained on these captions shows an $R^2$ uplift of 93.7\% and shrinks the mean absolute error (MAE) by 30\% compared to the baseline model. Qualitative analysis shows that high-impact tokens align with high-demand zones, offering lightweight support for heat planning in data-scarce regions.
\end{abstract}

\section{Introduction}
Minimizing dependency on fossil fuels is a crucial step towards fighting climate change. Space heating accounts for a larger share of energy use in both the commercial and residential sectors, making it a key focus for decarbonization efforts. In Europe, the policy goal is to optimize energy supply and use to reduce greenhouse gas emissions by 40\% by 2030 and 80\% by 2050~\cite{malcher2024strategies, boldrini2022role}. To achieve this, reducing heat demand through efficient upgrades to district heating systems and building retrofits is recognized as an effective strategy, given the carbon intensity of the current energy mix~\cite{novosel2020heat}. Assessing heat demand across the city is a crucial factor in selecting a decarbonization strategy. Certain high-precision heat demand evaluations depend on comprehensive geographic and energy datasets.

Nevertheless, key building attributes for energy consumption, such as construction year, geometry, insulation level, and heating system, are frequently incomplete, outdated, or restricted by data-protection regulations. To overcome these data gaps, prior work has relied on either bottom-up or top-down heat-mapping approaches~\cite{novosel2020heat}. Their accuracy, however, is constrained by the availability of comprehensive building metadata.

\begin{figure}[!t]
    \centering
    \includegraphics[width=\columnwidth]{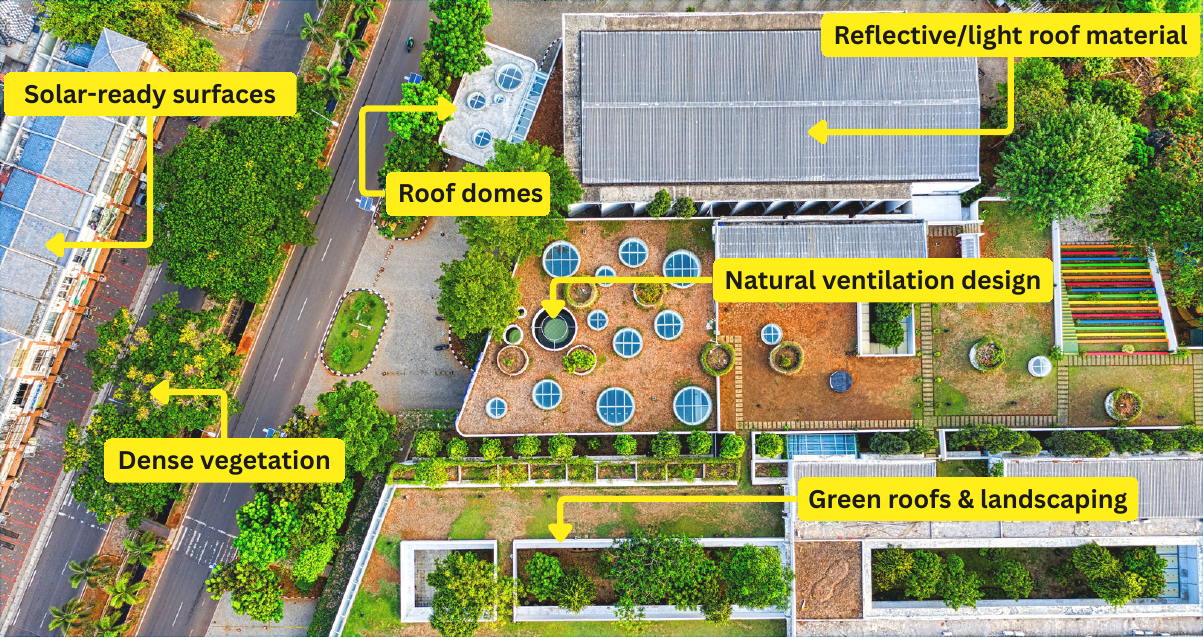}%
    \caption{Energy-relevant features such as solar-ready surfaces, green roof areas, reflective roofing, and surrounding vegetation that support heat demand estimation.}
    \label{fig:energy-features}
\end{figure}

Therefore, researchers have pivoted to publicly available remote sensing satellite imagery, which has become a foundation for urban analytics, from land‐use classification and building‐footprint extraction to population estimation. However, converting pixel data into quantitative energy metrics continues to present significant challenges~\cite{xu2019advanced,patino2013review,wang2022unetformer}. As a proxy, other features, such as roof material, building density, and vegetation presence, derived from images as shown in Fig.~\ref{fig:energy-features} can be used to estimate heat demand.

To bridge this gap, we propose HeatPrompt, a zero-shot pipeline that extracts energy-relevant semantic features from satellite images and uses them to estimate continuous annual heat demand. Our main contributions include the following: 

\begin{itemize}
    \item \textbf{Zero‐Shot Semantic Heat Mapping:}  
We introduce HeatPrompt, the first framework to model annual heat demand for municipal zones using RGB satellite images, binary isoline masks, and zero-shot semantic captions from a VLM, eliminating the need for manual labeling of energy-level features.
\item \textbf{Interpretable Pipeline:}
HeatPrompt not only outperforms Convolutional Neural Networks (CNN)‐based regressors but also highlights human‐readable semantic features that reveal the key drivers of heat demand, providing planners with transparent, actionable insights rather than opaque black‐box predictions.
\item \textbf{Reproducible Benchmark:}  
We provide open‐source code \footnote{\url{https://github.com/kundanthota/HeatPrompt}}
for data preparation, stratified cross‐validation, and paired \(t\)‐tests on absolute errors, establishing the first vision-language benchmark for urban heat‐demand estimation.
\end{itemize}

The paper is organized as follows: Section~\ref{sec:relatedwork} reviews related work on energy demand models and remote sensing data. Section~\ref{sec:method} details the data acquisition and the proposed method. Section~\ref{sec:results} shows the experimental results and an ablation study. In contrast, Section~\ref{sec:conclusion} concludes the paper and outlines the direction of future research.

\section{Related Work}
\label{sec:relatedwork}

\subsection{Heat Demand Mapping and Simulation Approaches}
Heat demand mapping serves as a basis for municipal heat transition planning, and traditional approaches fall into two categories: bottom-up and top-down models. Bottom-up models estimate thermal energy needs by aggregating detailed building-level features such as floor area, usage type, construction year, and insulation levels. For instance, Meha et al.~\cite{meha2020bottom} applied a bottom-up model for computing heating demand at the building level using geometric and usage parameters, then aggregating the results to a 100×100 m resolution map. These methods offer high spatial accuracy but are constrained by their reliance on comprehensive local data, which is often missing or outdated.

Top-down models address data scarcity by disaggregating aggregate heat consumption to finer scales using features such as land-use, population density, or floor space distribution. Novosel et al.~\cite{novosel2020heat} demonstrated such a top-down mapping methodology for data-poor areas by calibrating national consumption data with minimal local information. Recent strategic studies and national data reports highlight that accurate spatial heat demand information is key to cost-efficient, optimised district heating networks.~\cite{malcher2024strategies, lerbinger2023optimal, bundesamt2023energieverbrauch}

\subsection{Machine Learning for Energy Demand Prediction}

Machine learning techniques are increasingly used to estimate energy consumption and heating demand, particularly when data is missing in simulation models. Early works explored regression techniques and time-series models such as SARIMA to predict heat loads~\cite{fang2016evaluation}. More recent studies utilize deep neural networks (DNNs), Long Short-Term Memory (LSTMs), and CNNs for short-term load forecasting~\cite{frison2024evaluating}, leveraging meteorological and historical sensor data.

While these methods perform well, they often depend on historical data, limiting their use for estimating energy consumption in unmeasured buildings. To address this, Gao et al.~\cite{gao2020deep} and Morteza et al.~\cite{morteza2023deep} applied transfer learning to estimate building energy consumption in data-scarce settings. However, these techniques require at least partial building-level features such as insulation levels or age.

\subsection{Remote Sensing and VLMs in Urban Analytics}

Remote sensing has been used for urban classification, land cover analysis, and socio-economic inference~\cite{patino2013review, gao2020deep}. High-resolution aerial and satellite images are now routinely used to identify building footprints, roof types, and vegetation coverage. Wang et al.~\cite{wang2022unetformer} leverage UNetFormer for efficient semantic segmentation, enabling pixel-level information on rooftops, infrastructure, and urban context, which is used for various energy-relevant applications. 

More recently, foundation models such as OpenAI's Contrastive Language-Image Pre-training (CLIP)~\cite{radford2021learning} have enabled powerful vision-language alignment via contrastive pretraining. CLIP and its geospatial adaptations can perform zero-shot classification of satellite images by matching them against textual descriptions, eliminating the need for task-specific labels. RemoteCLIP, trained on a large corpus of remote sensing image-text pairs, significantly outperforms generic VLMs in scene understanding tasks and enables the extraction of rich semantic cues from RGB imagery~\cite{li2023rs, liu2024remoteclip, lin2025rs}. Instruction-tuned VLMs such as Large Language and Vision Assistant (LLaVA)~\cite{liu2023visual} and multimodal LLMs like GPT-4o~\cite{islam2024gpt}, Qwen2.5-VL~\cite{Qwen2.5-VL} push this further by enabling expert-like reasoning over images through tailored prompts.

Despite their proven effectiveness in remote sensing and general vision tasks, these models have not been widely applied in energy modeling. Our research addresses this gap by explaining that features extracted from VLMs, such as roof condition, building age, and urban density, meaningfully correlate with various energy-related parameters. These insights serve as a semantic proxy, enriching estimation processes by providing valuable information that is typically difficult to quantify without extensive field surveys. By leveraging these inferred features, we enhance traditional models, particularly in urban areas characterized by diverse structures and visual characteristics. 

\section{Proposed Method}
\label{sec:method}
\begin{figure*}[!t]
\centering
\includegraphics[width=\textwidth]{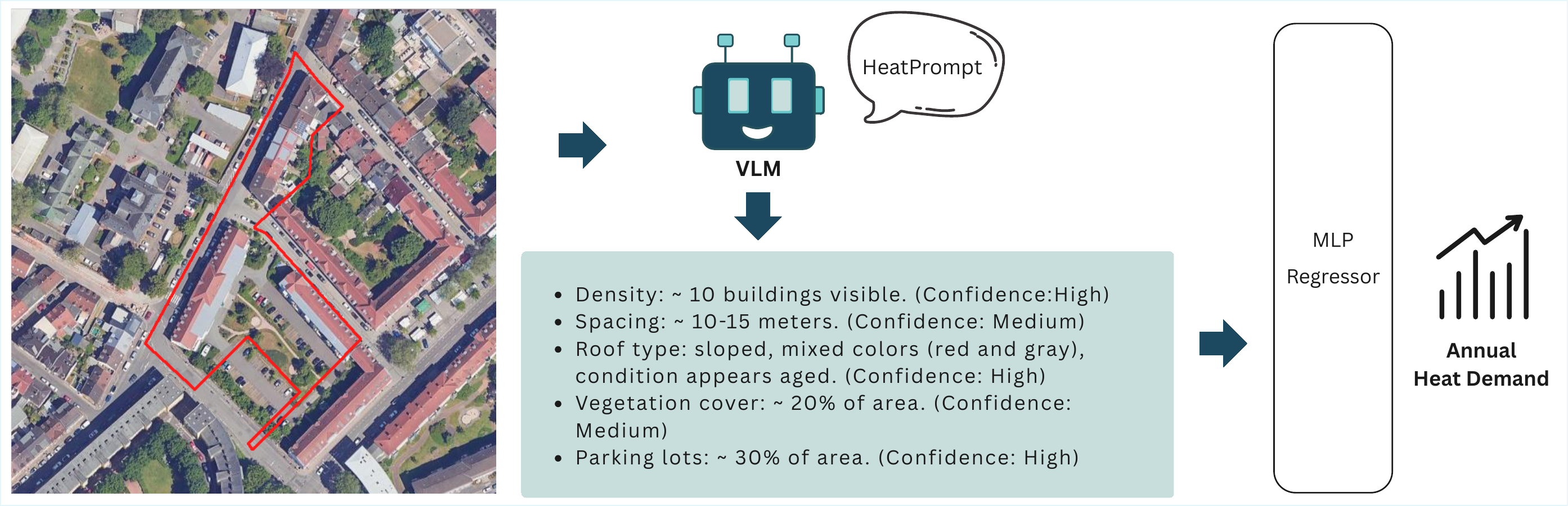}
\caption{Illustration of HeatPrompt's semantic feature extraction. An isoline mask (red boundary) is first overlaid on the 512×512 px satellite image, defining the region of interest. Then, a VLM is prompted to extract visual attributes and associated confidence scores from this RGBA composite. These semantic features are then used for heat‐demand regression.}
\label{fig:architecture}
\end{figure*}

\subsection{Dataset Acquisition}
\label{sec:dataset}
Satellite images are captured using the Esri World Imagery basemap~\cite{esri_arcgis_world_imagery}, and municipal heat planning isolines are overlaid onto these images as shown in Fig.~\ref{fig:architecture}. The objective of this study is to predict the annual heat demand based on the resulting satellite imagery.

Accordingly, each sample consists of an RGB image and a corresponding binary mask representing the region of interest, defined by a multipolygon isoline. As an example data source, we use the Rheinland-Palatinate (RLP) Energy Atlas~\cite{rlp2025energieatlas}, which provides municipal heat consumption isolines as geospatial multipolygons along with associated metadata, including the annual heat demand $y_i$ (MWh/a), planimetric area $A_i$ (m$^2$), and perimeter length $\ell_i$ (m).

Each polygon $P_i$ is reprojected to the Web Mercator coordinate reference system (EPSG:3857), following standard geospatial preprocessing practices~\cite{sturm2020some}. We compute its centroid and define a square sampling window of side
 
\[
  s_i \;=\; 1.05\,\max\!\{\text{width}(P_i),\,\text{height}(P_i)\}
\]
to introduce a 5\% contextual margin and capture a 512×512 px RGB satellite Image $I_i$, ensuring sub‐meter ground sampling distance.  

To isolate the region of interest, we rasterize $P_i$ at the exact resolution, producing a binary mask $M_i\in\{0,1\}^{512\times512}$ where $M_i(x,y)=1$ only if the pixel lies inside the isoline. The resulting dataset consists of 1,677 samples with aligned triplets ${I_i, M_i, y_i}$, along with per-polygon GIS attributes $(A_i, \ell_i)$, and forms the foundation for feature extraction and regression. In addition to polygon-level metadata, we enrich the region of interest with detailed building-level footprints extracted from OpenStreetMap (OSM) such as building types (e.g: Single Family House (SFH), Multi Family House (MFH), Terraced Row House (TR), Non-Residential (NR))~\cite{mooney2017review}, Census 2011~\cite{Destatis_Zensus2011}, supplemented by LOD2 building height wireframes ~\cite{BKG_LoD2_DE_2025}, as shown in the Algorithm~\ref{alg:orchestrator}. By mapping these features with each isoline, we construct a high-resolution dataset that serves as both input and ground truth for model validation.

\begin{algorithm}[!b]
\caption{Construction of Aligned Heat Demand Dataset with Auxiliary Attributes}
\label{alg:orchestrator}
\centering
\begin{minipage}{0.95\columnwidth}
\raggedright\small
\begin{spacing}{1.25}
\begin{algorithmic}
\State \textbf{Input:} Polygons $\{P_i\}$ from Energieatlas with metadata $y_i, A_i, \ell_i$
\State \textbf{Output:} Triplets $\{I_i, M_i, y_i\}$ with auxiliary building-level attributes

\ForAll{$P_i$}
    \State Reproject $P_i \rightarrow$ EPSG:3857
    \State Define sampling window of side $s_i = 1.05 \cdot \max\{\text{width}(P_i),\,\text{height}(P_i)\}$
    \State Fetch RGB satellite image $I_i \in \mathbb{R}^{512 \times 512 \times 3}$
    \State Rasterize binary mask $M_i \in \{0,1\}^{512 \times 512}$ from $P_i$
\EndFor

\ForAll{buildings $B_j \subseteq P_i$}
    \State Extract OSM tags $\rightarrow$ estimate use-type $t_j \in \{\text{SFH}, \text{TH}, \text{MFH}, \text{NR}\}$
    \State Link LOD2 roof wireframe $\rightarrow$ compute height $h_j$
    \State Nearest-neighbor match to census data $\rightarrow$ age class $e_j$
    \State Compute floor area $f_j = \text{area}(B_j) \cdot \left\lfloor h_j / h_{\text{floor}} \right\rfloor$
\EndFor

\ForAll{$P_i$}
    \State Append derived features $(A_i, \ell_i)$ to $\{I_i, M_i, y_i\}$
\EndFor

\State \Return Full dataset $\{I_i, M_i, y_i, A_i, \ell_i, \text{building features}\}$

\end{algorithmic}
\end{spacing}
\end{minipage}
\end{algorithm}

\subsection{HeatPrompt: Semantic Feature Extraction}
\label{sec:feature_extraction}

For each 512×512 satellite image \(I_i\) and its aligned binary mask \(M_i\), we construct an RGBA composite by inserting \(M_i\) into the alpha channel, thereby highlighting the region of interest. This composite image is fed to a VLM under a domain‐specific prompt that instructs the model to act as a municipal heat planner, as shown in Fig.~\ref{fig:architecture}, and list the five most salient visual factors affecting heating demand. The resulting output is transformed into a fixed‐length vector \(z_i^{\rm sem}\in\mathbb{R}^{512}\) using Nomic text-embeddings~\cite{nussbaum2024nomic}. These semantic embeddings capture human‐interpretable features, such as building density, roof characteristics, and vegetation cover, along with the prediction confidence level, that can be fed into a regressor without any additional fine‐tuning or pixel‐level supervision. 

\subsection{Regression on Semantic Embeddings}

Let $\mathbf{x}_i^{(0)} = [A_i,\;\ell_i] \in \mathbb{R}^2$ denote the GIS attributes from the RLP Energy Atlas, and let $z_i^{\mathrm{sem}} \in \mathbb{R}^{512}$ be the semantic embedding derived from the VLM captions. Additionally, let $\mathbf{x}_i^{\mathrm{comp}} \in \mathbb{R}^d$ represent the building composition features derived from LOD2 geometry, OpenStreetMap tags, and census metadata. These features include aggregated descriptors such as height, floor count, floor area, construction decade, and building type ratios per isoline region.

We define the full feature vector as:
\[
  \mathbf{x}_i^{\mathrm{full}}
  = 
  \begin{bmatrix}
    \mathbf{x}_i^{(0)} \\[0.3em]
    z_i^{\mathrm{sem}} \\[0.3em]
    \mathbf{x}_i^{\mathrm{comp}}
  \end{bmatrix}
  \in \mathbb{R}^{514 + d},
\]
where $d$ is the dimensionality of the building composition vector (e.g., $d \approx 11$ after feature aggregation).

We train and evaluate a Multi-Layer Perceptron (MLP) with two hidden layers of 64 units each, a dropout rate of 0.2, and optimized using a learning rate of $1\times10^{-4}$ with weight decay for up to 250 epochs with early stopping. The model is trained on the dataset
\[
X_{\mathrm{full}} = \{ \mathbf{x}_i^{\mathrm{full}} \}_{i=1}^N,
\]
using stratified five-fold cross-validation (CV) based on heat-demand quintiles to ensure that each fold preserves the overall distribution of target values.

For each fold, we report the coefficient of determination ($R^2$) and Mean Absolute Error (MAE).

\section{Results and Discussion}
\label{sec:results}

\begin{figure}[b]
 \centering
 \includegraphics[width=1\columnwidth]{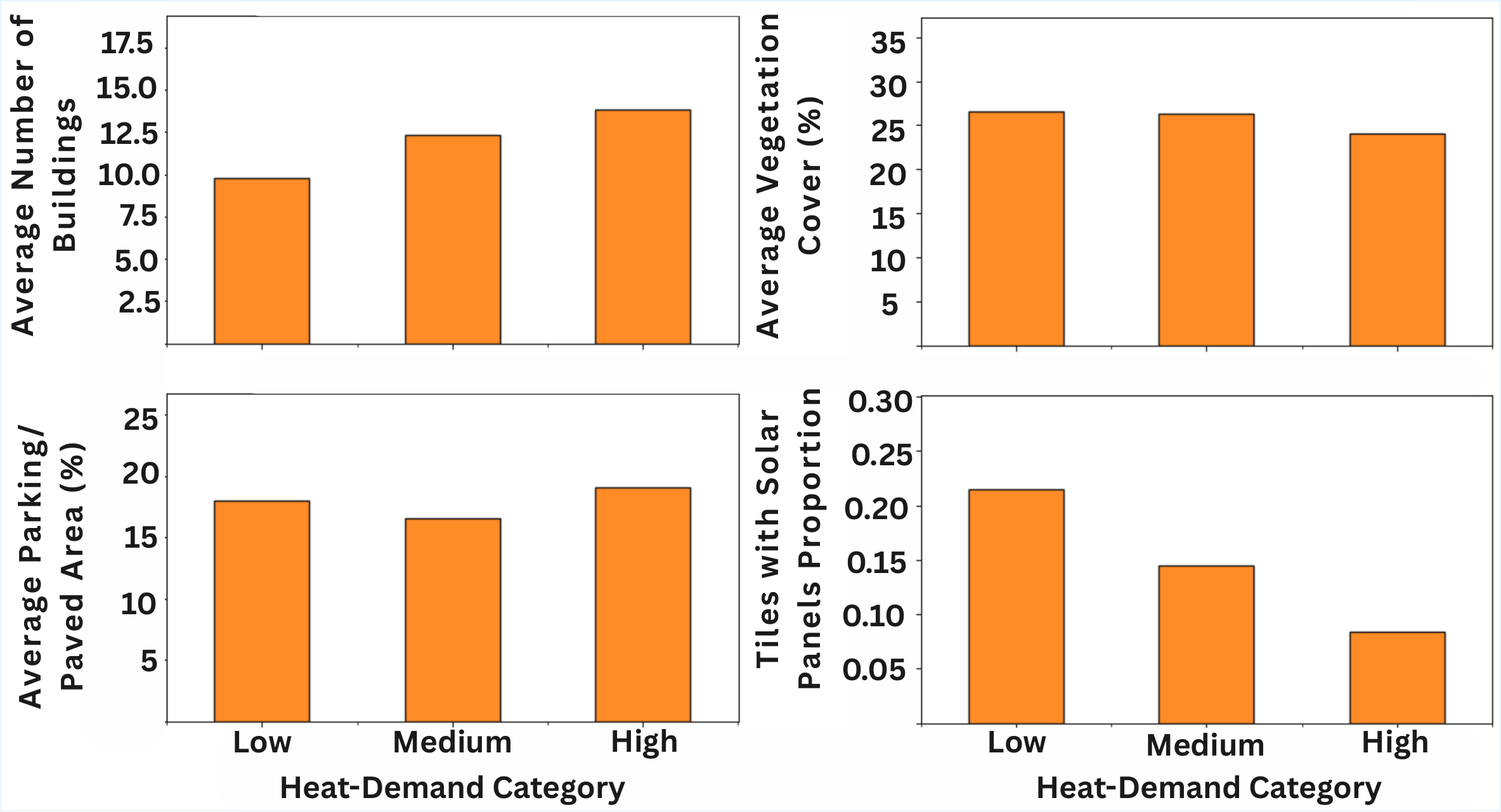}
 \caption{Semantic trends across heat demand terciles. Visual signals, such as tree cover and roof condition, correlate with modeled demand.}
 \label{fig:semantic_trends}
 \end{figure}
 
\subsection{Structured Feature Regression Performance}
Table~\ref{tab:regressor_results} compares the performance of various regression models trained on structured building attributes using stratified 5-fold cross-validation. The bottom-up simulation baseline, based on physical energy modeling, achieved an average $R^2$ of 0.32 and MAE of 287.1, reflecting its limited predictive capacity in the absence of fine-grained or updated building information.
Among traditional ML models, Ridge regression performed best ($R^2$=0.47), with a 46.9\% improvement over the baseline. Tree-based methods such as Random Forests and Gradient Boosting yielded marginal improvements.
A simple MLP model achieved the highest performance, with an average $R^2$ of 0.51 and an MAE of 239.0, resulting in a 59.4\% increase over the simulation baseline, making it the base regressor for further ablation and feature expansion.
\begin{table}[t]
 \centering
 \footnotesize
 \setlength{\tabcolsep}{4pt}
 \begin{tabular}{|l|c|c|c|}
 \hline
 \textbf{Model} & \textbf{$R^2$(± std)} & \textbf{MAE (± std)} & \textbf{\% $R^2$} \\
 \hline
 Baseline simulation & 0.32 ± 0.30 & 287.1 ± 56.4 & -- \\
 Linear Regression~\cite{su2012linear} & 0.43 ± 0.15 & 244.7 ± 39.3 & +34.4\% \\
 Ridge Regression~\cite{mcdonald2009ridge} & 0.47 ± 0.15 & 240.9 ± 38.8 & +46.9\% \\
 Random Forest~\cite{rigatti2017random} & 0.35 ± 0.12 & 238.1 ± 42.6 & +8.6\% \\
 HistGradBoost~\cite{guryanov2019histogram} & 0.34 ± 0.41 & 237.8 ± 49.0 & +6.3\% \\
 MLP~\cite{popescu2009multilayer} & \textbf{0.51 ± 0.17} & \textbf{239.0 ± 37.7} & \textbf{+59.4\%} \\
 \hline
 \end{tabular}
 \caption{\textbf{Baseline Regression Performance Using Structured Building Features.}
Comparison of linear, tree-based, and neural regression models evaluated with 5-fold cross-validation.}

 \label{tab:regressor_results}
 \end{table}

\subsection{Impact of Semantic Features from Vision Models}
To evaluate the contribution of visual semantics, we enrich the structured features with embeddings from various pretrained vision-language and CNN backbones. Table~\ref{tab:semantic_feature_comparison} reports the resulting performance. Each model uses the same MLP architecture and is evaluated via stratified 5-fold cross-validation.
Across all configurations, augmenting the simulation baseline with latent visual features leads to substantial gains. GPT-4o-derived text embeddings show the most improvement, with an $R^2$ of 0.62 and MAE of 200.7, marking a 93.7\% relative gain over the bottom-up simulation. CLIP also achieves competitive performance, strengthening the idea that semantically rich features can serve as proxies for metadata that can't be manually extracted, which are energy-oriented.
\begin{table}[t]
 \centering
 \footnotesize
 \setlength{\tabcolsep}{5pt}
 \begin{tabular}{|l|c|c|c|}
 \hline
 \textbf{Features} & \textbf{$R^2$} & \textbf{MAE} & \textbf{\% $R^2$} \\
 \hline
 Baseline simulation & 0.32 ± 0.30 & 287.0 ± 56.4 & -- \\
 + ResNet50~\cite{mascarenhas2021comparison} & 0.59 ± 0.18  & 205.3 ± 27.3 & +84.4\% \\
 + EfficientNet-B0~\cite{tan2019efficientnet} & 0.58 ± 0.17 & 208.6 ± 32.6 & +81.3\% \\
 + DenseNet121~\cite{huang2017densely} & 0.57 ± 0.18 & 213.3 ± 38.4  & +78.1\% \\
 + CLIP~\cite{radford2021learning} & 0.60 ± 0.18 & 205.4 ± 22.3 & +87.5\% \\
 + Qwen2.5-VL~\cite{Qwen2.5-VL} & 0.58 ± 0.22 & 210.4 ± 27.1 & +81.3\% \\
 + GPT-4o~\cite{islam2024gpt} & \textbf{0.62 ± 0.17} & \textbf{ 200.7 ± 33.2 } & \textbf{+93.7\%} \\
 \hline
 \end{tabular}
 \caption{\textbf{Impact of Semantic Vision Features on Energy Demand Prediction Performance.}
Ablation study comparing baseline simulation with vision-augmented models using 5-fold cross-validation.}

 \label{tab:semantic_feature_comparison}
 \end{table}

\subsection{Semantic Trends and Visual Interpretability}
In addition to numerical gains, the semantic features exhibit interpretable relationships with known energy drivers. As shown in Fig.~\ref{fig:semantic_trends}, areas with high predicted heat demand exhibit lower vegetation presence, fewer solar panels, and higher structural density, all of which are visually inferred via GPT-4o.

\begin{figure}[t]
 \centering
 \begin{subfigure}[b]{0.23\textwidth}
 \includegraphics[width=\textwidth]{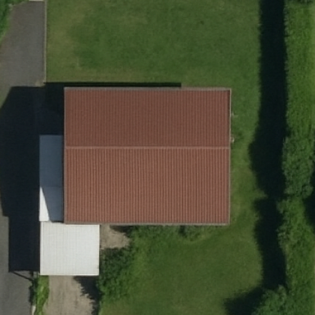}
 \caption*{\textbf{New Roof}\\High vegetation cover\\Estimated Demand:\\ \textbf{100~kWh/m\textsuperscript{2}}}
 \end{subfigure}
 \hfill
 \begin{subfigure}[b]{0.23\textwidth}
 \includegraphics[width=\textwidth]{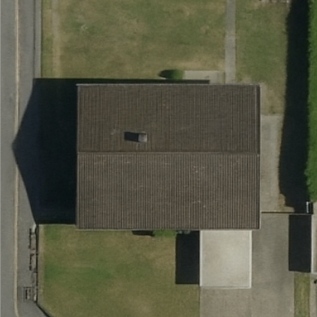}
 \caption*{\textbf{Old Roof (tile)}\\Low vegetation cover\\Estimated Demand:\\ \textbf{150~kWh/m\textsuperscript{2}}}
 \end{subfigure}
 \caption{Visual comparison of similar buildings showing semantic cues influencing heat demand.}
 \label{fig:vlm_annotated}
 \end{figure}
 
\subsection{Qualitative Comparison: Role of Visual Cues}
Fig.~\ref{fig:vlm_annotated} provides a qualitative comparison of two geometrically similar buildings. Despite having similar size and shape, they differ significantly in visual characteristics: one has a modern roof with high greenery, while the other has an old tiled roof and bare surroundings. Traditional simulation models would treat both identically. In contrast, our approach, through zero-shot VLM prompting, identifies such cues and yields distinctly different heat demand predictions, reflecting more realistic outcomes.

These results strengthen the value of using vision-based semantic features in such pipelines, making it easier and more automated for energy planners to consider as many factors as possible that directly correlate with energy demand. Meanwhile, embeddings such as CLIP achieve notable results, making them a practical substitute for many physical variables, such as roof condition and land cover, and even for masking features for data protection.

\section{Conclusion}
\label{sec:conclusion}

We introduce HeatPrompt, a zero-shot vision-language framework that uses GPT-4o-generated captions along with basic GIS and building-level features to estimate annual heat demand from satellite images. This addition of semantic features yields a consistent increase in the explained variance \(R^2\) by 93.7\% from $(0.32\pm0.30)$ to $(0.62\pm0.17)$ and cuts down the MAE by 30\%. Unlike CNN-based regressors, HeatPrompt also produces human-readable tokens, such as building density, roof type, and vegetation cover, thereby offering a user-friendly energy modeling pipeline.

Looking forward, we will develop energy-aware visual-language embeddings, inspired by CLIP and tailored to urban-energy semantics. This will enable planners to directly retrieve attributes such as roof age, insulation properties, PV coverage, and vegetation shielding from satellite images, as well as indirect proxies for less observable factors like insulation quality. These embeddings will capture latent correlations among building morphology, construction era, and neighborhood context that influence thermal performance.
These vectors can be integrated into energy demand models without task-specific retraining. In parallel, we will explore lightweight fine-tuning of promptable VLMs to focus the embedding space on energy-relevant factors. Together, these efforts aim to provide a reusable and interpretable embedding layer for data-light energy modeling that planners can query, visualize, and use.

\section*{Acknowledgments}
This work was carried out within the project "New Data for the Energy Transition" (NEED)~\cite{Duchon24-platform-ecosystem} and was supported by BMWE under Grant No. 03EN3077J. Portions of the manuscript text (e.g., wording suggestions and Grammar) were drafted with ChatGPT's Assistance. The authors reviewed, edited, and take full responsibility for all content.

\balance
\bibliographystyle{unsrt}
\bibliography{bibliography}

\end{document}